\documentclass[11pt,a4paper]{article}
\usepackage{emnlp2018}
\usepackage{times}
\usepackage{latexsym}
\usepackage{tikz}

\usepackage{url}
\usepackage{rotate}
\aclfinalcopy 

\usepackage{graphicx}
\usepackage[font=small,skip=5pt]{caption}
\usepackage{subcaption}
\usepackage{amssymb}
\usepackage{amsfonts}
\usepackage{amsbsy}
\usepackage[safe]{tipa}
\usepackage{float}
\usepackage{xr} 

\usepackage{url}

\usepackage{amssymb}
\usepackage{gb4e}
\noautomath
\usepackage{tikz}
\usepackage{pgfplots}
\usepackage{amsbsy}
\usepackage{amssymb}
\usetikzlibrary{positioning}
\usetikzlibrary{calc}
\usetikzlibrary{fit}
\usetikzlibrary{patterns,decorations.pathreplacing,decorations.markings,arrows}
\usetikzlibrary{shapes.geometric}

\tikzset{->-/.style={decoration={
  markings,
  mark=at position 0.65 with {\arrow[scale=1.5,#1]{>}}},postaction={decorate}}}
\tikzset{dashdot/.style={dash pattern=on .4pt off 3pt on 4pt off 3pt}} 

\usepackage{rotating}

\newcommand{\real}{\mathbb{R}} 
\newcommand{\Prob}{\mathbb{P}} 

\newcommand{\bfC}{\textbf{C}}
\newcommand{\bfa}{\textbf{a}}
\newcommand{\bfb}{\textbf{b}}
\newcommand{\bfc}{\textbf{c}}

\newcommand{\bfw}{\mathbf{w}}

\newcommand{\hatbfc}{\widehat{\textbf{c}}}
\newcommand{\bfepsilon}{\boldsymbol{\epsilon}}

\newcommand{\sfx}{\mathsf{\small x}}
\newcommand{\sfy}{\mathsf{\small y}}
\newcommand{\sfz}{\mathsf{\small z}}

\newcommand{\hatsfx}{\widehat{\textsf{\small x}}}
\newcommand{\hatsfy}{\widehat{\textsf{\small y}}}
\newcommand{\hatsfz}{\widehat{\textsf{\small z}}}

\newcommand{\hatm}{\widehat{m}}

\newtheorem{ccorollary}{\bf Corollary}

\newtheorem{lLemma}{\bf Lemma} 
\newenvironment{Lemma}
{\begin{lLemma}}{\end{lLemma}}

\newtheorem{pProposition}{\textbf{Proposition}}                
\newenvironment{Proposition}
{\begin{pProposition}}{\end{pProposition}}

\newtheorem{ddefinition}{\textbf{Definition}}

\title{Equiprobable mappings in weighted constraint grammars}

\author{Arto Anttila \\
  Stanford University \\
  {\tt\small anttila@stanford.edu} \\\And
  Scott Borgeson \\
  Stanford University \\
  {\tt\small borgeson@stanford.edu} \\\And
  Giorgio Magri \\
  CNRS \\
  {\tt\small magrigrg@gmail.com}
  }

\date{}

\begin{document}
\maketitle
\begin{abstract}
We show that MaxEnt is so rich that it can distinguish between any two different mappings: there always exists a nonnegative weight vector which assigns them different MaxEnt probabilities. Stochastic HG instead does admit equiprobable mappings and we give a complete formal characterization of them. We compare these different predictions of the two frameworks on a test case of Finnish stress. 
\end{abstract}

\section{Introduction}

This paper compares two frameworks for probabilistic constraint-based phonology: {\em Stochastic Harmonic Grammar} (SHG; Boersma and Pater, 2016\nocite{BoersmaPater(2016)})\footnote{
Boersma and Pater \shortcite{BoersmaPater(2016)} actually use the term ``noisy HG'' instead of ``stochastic HG''. We prefer ``stochastic HG'' to stress the complete analogy with Boersma's \shortcite{Boersma(1997),Boersma(1998)} earlier framework of stochastic OT. Furthermore, we prefer to use ``stochastic'' to describe a property of the framework, reserving ``noisy'' to describe a property of the learning scenario (as opposed to noise-free). 
} and {\em Maximum Entropy} (ME; Goldwater and Johnson, 2003; Hayes and Wilson, 2008\nocite{GoldwaterJohnson(2003),HayesWilson(2008)}). Recent literature has documented a few realistic quantitative patterns which seem to admit a better fit in ME than in SHG \cite{SmithPater(2017),ZurawHayes(2017),Hayes(2017)}. These findings suggest that ME is a richer probabilistic framework than SHG (relative to the same constraint set). But how much richer? Can these anecdotal observations reported in the literature be systematized into a principled formal comparison between SHG and ME probabilistic typologies? This paper is part of a larger project trying to address this question. In particular, this paper compares ME and SHG from the perspective of their {\em equiprobable mappings}. That is phonological mappings which are always assigned the same probability and are therefore phonologically equivalent despite being distinguished by the constraint set.




Section \ref{section: phonological equivalence} motivates this notion of equiprobability within phonological theory. Section \ref{section: A new result on uniform probability identities in SHG and ME} then shows that the ME typology is so rich that it admits no equiprobable mappings: for any two mappings distinguished by the constraints, there exists an ME grammar that distinguishes between them, namely assigns them different probabilities. This typological richness is peculiar to ME and does {\em not} extend to other implementations of probabilistic constraint-based phonology such as SHG. Indeed, Section \ref{section: SHG allows for equiprobable mappings} shows that the equiprobable SHG mappings are exactly those mappings which are indistinguishable by categorical {\em Harmonic Grammars} (HG; Legendre {\em et al.}, 1990a,b; Smolensky and Legendre, 2006\nocite{LegendreMiyataSmolenskyFoundations(1990),LegendreMiyataSmolenskyApplication(1990),SmolenskyLegendre(2006)}) and thus provides a complete characterization of SHG equiprobability.


These formal results are presented informally. A detailed proof of the ME result is provided in a final appendix. The proof of the SHG result is analogous and it is omitted for reasons of space (see the longer version of this paper available on the authors' websites). Our discussion rests on some earlier results on uniform SHG and ME probability inequalities from Anttila and Magri \shortcite{AnttilaMagri(2018AMP)}, recalled in Section \ref{section: Background}. 


Is the richness of ME relative to SHG typologies an empirical advantage or a case of unmotivated  overgeneration? Section \ref{section: Equiprobable mappings in Finnish stress} provides some preliminary evidence that the latter might be the case, by looking at the case of Finnish stress. We compute SHG equiprobable mappings using the formal characterization obtained in Section \ref{section: SHG allows for equiprobable mappings}. We show that a large corpus of Finnish provides preliminary empirical support for these mappings indeed being equiprobable. Finally, we show that ME breaks up these equiprobabilities in a way that is phonologically counterintuitive.

\section{Equiprobability}
\label{section: phonological equivalence}

A typical phonological process applies uniformly to all forms that share some relevant property, but ignores the irrelevant ways in which they differ. For example, in Latin, stress targets heavy syllables, but ignores vowel quality; in English, aspiration targets voiceless stops, but ignores place of articulation; in Finnish, vowel harmony targets [$\pm$back], but ignores the number of syllables. This means that words with the same distribution of heavy and light syllables are stressed alike; voiceless stops are aspirated alike; and words of any length harmonize alike. These phonological {\em equivalences} are a key property of phonological systems.


Derivational phonology captures these equivalences straightforwardly: phonological rules are allowed to refer to only the shared property that defines a natural class, ignoring everything else. To illustrate, the Finnish vowel harmony rule can be simply written as $\mbox{V} \rightarrow [\alpha\mbox{back}] / \mbox{V}[\alpha\mbox{back}]\mbox{C}_0\_\!\_$. This rule directly encodes the fact that harmony targets [$\pm$back] but ignores any other properties such as, say, the number of syllables. Thus, the monosyllabic \textsf{\small /maa/} `country' and the disyllabic \textsf{\small /kaava/} `formula' trigger back harmony on the suffix \textsf{\small /-n\"{a}/} `{\sc essive\/}' in exactly the same way. In other words, they are equivalent for vowel harmony.


The situation is {\em prima facie} less obvious in constraint-based phonology. A candidate may contain multiple constraint violations, some relevant, some irrelevant, but all simultaneously visible and potentially interacting. Yet, categorical implementations of constraint-based phonology are well known to readily predict these desired phonological equivalences. To illustrate, consider an HG grammar for Finnish vowel harmony based on the constraints in Table \ref{vowel-harmony-constraints}, from \citet{RingenHeinamaki(1999)}. The back harmony mappings $\textsf{\small /maa-n\"{a}/} \rightarrow \textsf{\small [maana]}$ and $\textsf{\small /kaava-n\"{a}/} \rightarrow \textsf{\small [kaavana]}$ can be shown to be HG equivalent: no matter the weighting, no HG grammar succeeds on one but fails on the other.


How should phonological equivalence be extended from the categorical to the probabilistic setting? We submit that equiprobability provides an answer to this question. In fact, let us recall that a {\em probabilistic phonological grammar} is a function which assigns to each underlying representation (UR) $\sfx$ a probability distribution $\Prob(\sfy \,|\, \sfx)$ over the corresponding set of candidate surface representations (SRs) $\sfy$. We consider two mappings $(\sfx, \sfy)$ and $(\hatsfx, \hatsfy)$ of the two URs $\sfx, \hatsfx$ to the two SRs $\sfy, \hatsfy$. We say that these two mappings are {\em (uniformly) equiprobable} provided there is no probabilistic grammar in the typology considered which assigns a different probability to those two mappings, namely such that $\Prob(\sfy \,|\, \sfx) \not= \Prob(\hatsfy \,|\, \hatsfx)$. To illustrate, the equivalence between the two mappings $\textsf{\small /maa-n\"{a}/} \rightarrow \textsf{\small [maana]}$ and $\textsf{\small /kaava-n\"{a}/} \rightarrow \textsf{\small [kaavana]}$ is captured in a probabilistic setting through the requirement that their probabilities $\Prob(\textsf{\small [maana]} \,|\,\textsf{\small /maa-n\"{a}/})$ and $\Prob( \textsf{\small [kaavana]} \,|\, \textsf{\small /kaava-n\"{a}/})$ always coincide. In other words, the probability of vowel harmony does not depend on the number of syllables.\footnote{
Note that this is quite different from the well-known case of Hungarian vowel harmony where suffixes show different degrees of back-front variation after stems with both back and neutral vowels depending on the number of neutral vowels; see, e.g., \citet{HayesLonde(2006)}, \citet{HayesEtAl(2009)}, and \citet{Zymet(2015)}. In our  Finnish example, all the stem vowels are unambiguously back, yet our Proposition \ref{Proposition: equiprobable mappings in ME} below says that ME fails to guarantee that the suffix harmony is invariably back.
}


\begin{table}
\hspace{-0.2cm}\renewcommand{\tabcolsep}{0.1cm}\begin{tabular}{rl}
{\sc *Int}[+back]: 	& No vowel between [+back] and 
\\
				& right word edge
\smallskip\\
{\sc *Int}[-back]: 	& No vowel between [-back] and 
\\				& right word edge
\smallskip\\
{\sc Ident-Root}: 	& Be faithful to \textsf{\small /a, \"{a}/} in roots
\smallskip\\
{\sc Ident}: 		& Be faithful to \textsf{\small /a, \"{a}/}
\smallskip\\
\end{tabular}
\caption{Constraints for Finnish vowel harmony}
\label{vowel-harmony-constraints}
\end{table}


As we will see in Section \ref{section: SHG allows for equiprobable mappings}, two mappings are equivalent according to categorical HG if and only if they are equiprobable in SHG. This result suggests that equiprobability is indeed the right extension of the notion of phonological equivalence from the categorical to the probabilistic setting. Surprisingly, we will see in Section \ref{section: A new result on uniform probability identities in SHG and ME} that ME instead allows for no equiprobable mappings and thus fails to capture the notion of phonological equivalence.

\section{Formal background}
\label{section: Background}

Our characterization of ME and SHG equiprobability in sections \ref{section: A new result on uniform probability identities in SHG and ME}-\ref{section: SHG allows for equiprobable mappings} rests on some results from Anttila and Magri (2018; A\&M) recalled here.


\paragraph{HG} 
A {\em weight vector} $\bfw = (w_1, \dots, w_n)$ assigns nonnegative weights $w_1, \dots, w_n \geq 0$ to $n$ underlying phonological constraints $C_1, \dots, C_n$. The phonological quality of a phonological mapping $(\sfx, \sfy)$ of a UR $\sfx$ and a candidate SR $\sfy$ is quantified by its {\em harmony} $H_{\bfw}(\sfx, \sfy)$. This quantity is defined as the weighted sum of the constraint violations multiplied by $-1$, namely $H_{\bfw}(\sfx, \sfy) = - \sum_{k=1}^n w_k C_k(\sfx, \sfy)$. Mappings with large harmony have small constraint violations. The HG grammar corresponding to  a weight vector $\bfw$ maps a UR $\sfx$ to the candidate SR $\sfy$ such that the mapping $(\sfx, \sfy)$ has a larger harmony than the mapping $(\sfx, \sfz)$ corresponding to any other candidate $\sfz$ of $\sfx$. In this case, we say that $\sfy$ is the {\em winner} while any other candidate $\sfz$ is a {\em loser}.


HG thus has an intrinsic comparative nature: absolute numbers of violations are irrelevant, what matters is only the comparison between the violations of the loser and those of the winner. To bring out this intuition, we define the {\em difference vector} $\bfC(\sfx, \sfy, \sfz)$ for a UR $\sfx$, an intended winner candidate $\sfy$, and an intended loser candidate $\sfz$ as in (\ref{ex: antecedent difference vector}). This vector has a component for each constraint $C_k$ defined as the difference between the number $C_k(\sfx, \sfz)$ of violations assigned by $C_k$ to the loser mapping $(\sfx, \sfz)$ minus the number $C_k(\sfx, \sfy)$ of violations assigned to the winner mapping $(\sfx, \sfy)$.
\begin{equation}
\label{ex: antecedent difference vector}
\bfC(\sfx, \sfy, \sfz) 
=
\left[ \begin{array}{c}
C_1(\sfx, \sfz) - C_1(\sfx, \sfy)
\\[-0.1cm]
\vdots
\\[-0.1cm]
C_k(\sfx, \sfz) - C_k(\sfx, \sfy)
\\[-0.1cm]
\vdots
\\[-0.1cm]
C_n(\sfx, \sfz) - C_n(\sfx, \sfy)
\end{array} \right]
\end{equation}
SHG and ME are two probabilistic extensions of this underlying categorical HG model.


\paragraph{SHG}
The SHG probability $\Prob_{\bfw}^{\mbox{\tiny SHG}}(\sfy \,|\, \sfx)$ that a UR $\sfx$ is mapped to a SR $\sfy$ according to the weight vector $\bfw$ is the probability of sampling $n$ numbers $\bfepsilon = (\epsilon_1, \dots, \epsilon_n)$ independently according to a distribution $\mathcal{D}$ in such a way that the HG grammar corresponding to the weight vector $\bfw + \bfepsilon = (w_1+\epsilon_1, \dots, w_n+\epsilon_n)$ indeed maps $\sfx$ to $\sfy$. A\&M prove the following Lemma \ref{Lemma: probability inequalities in SHG} about {\em uniform} probability inequalities in SHG, namely inequalities which hold for every choice of the weight vector. 

\begin{Lemma} 		\label{Lemma: probability inequalities in SHG}
Consider two mappings $(\sfx, \sfy)$ and $(\hatsfx, \hatsfy)$. Assume that the UR $\sfx$ comes with only a finite number $m$ of loser candidates $\sfz_1, \dots, \sfz_m$ (besides the winner candidate $\sfy$) and that the mapping $(\sfx, \sfy)$ is possible in HG (namely, $\sfy$ beats the losers $\sfz_1, \dots, \sfz_m$ relative to some nonnegative weight vector). The SHG probability inequality $\Prob_{\bfw}^{\mbox{\tiny SGH}}(\sfy \,|\, \sfx) \leq \Prob_{\bfw}^{\mbox{\tiny SGH}}(\hatsfy \,|\, \hatsfx)$ holds uniformly for every choice of the nonnegative weight vector $\bfw$ if and only if for every loser candidate $\hatsfz$ of the UR $\hatsfx$, there exist $m$ nonnegative coefficients $\lambda_1, \dots, \lambda_m \geq 0$ (one for each loser candidate $\sfz_1, \dots, \sfz_m$ of the UR $\sfx$) such that
\begin{equation}
\label{ex: Appendix; HG T-orders; equivalent condition}
\bfC(\hatsfx, \hatsfy, \hatsfz) \geq \sum_{i=1}^m \lambda_i \, \bfC(\sfx, \sfy, \sfz_i)
\end{equation}
namely the difference vector $\bfC(\hatsfx, \hatsfy, \hatsfz)$ is at least as large (constraint by constraint) as the sum of the difference vectors $\bfC(\sfx, \sfy, \sfz_i)$ each rescaled by a corresponding nonnegative coefficient $\lambda_i$.\footnote{
{\em The two assumptions made by the lemma----that the UR $\sfx$ comes with only a finite number of losers and that the mapping $(\sfx, \sfy)$ is possible in HG----are non-restrictive. In fact, if a mapping $(\sfx, \sfy)$ is impossible in HG, then its SHG probability $\Prob_{\bfw}^{\mbox{\tiny SGH}}(\sfy \,|\, \sfx)$ can be shown to be equal to zero for every choice of the nonnegative weight vector $\bfw$. The probability inequality $\Prob_{\bfw}^{\mbox{\tiny SGH}}(\sfy \,|\, \sfx) \leq \Prob_{\bfw}^{\mbox{\tiny SGH}}(\hatsfy \,|\, \hatsfx)$ thus holds uniformly, because its left hand side is always equal to zero. The assumption made by the lemma that the mapping $(\sfx, \sfy)$ is possible in HG is therefore non-restrictive. Furthermore, HG has the property that only a finite number of candidates of any given UR win according to some weights \cite{Magri(2019Finiteness)}. All other candidates are redundant because impossible no matter how the weights are chosen. Since HG impossible mappings have zero SHG probability, the candidate set of any underlying form can always be assumed to be finite without loss of generality in SHG. The assumption made by the lemma that the UR $\sfx$ comes with only a finite number of losers is therefore non-restrictive.}
}
\hfill{$\Box$}
\end{Lemma}

\begin{figure}[t]
\centering
\begin{tabular}{cc}
	\begin{tikzpicture}[scale=0.3]
	\draw [thin] (-4.5, 0) -- (7, 0);
	\draw [thin] (0, -1) -- (0, 8);
\node at (-2, 5) [inner sep=0pt, label=left:{\footnotesize $\sfz_1$}] {$\bullet$} ;
\node at (0, 1) [inner sep=0pt, label=left:{\footnotesize $\sfz_2$}] {$\bullet$};
\node at (5, 2) [inner sep=0pt, label=right:{\footnotesize $\sfz_3$}] {$\bullet$};
\node at (3, 4) [inner sep=0pt, label=right:{\footnotesize $\sfz_4$}] {$\bullet$};
	\draw [dashed] (0, 0) -- (-2.4, 6); 
	\draw [dashed] (0, 0) -- (5.5, 2.2); 
	\fill [pattern=north west lines, pattern color=gray] (0, 0) to (-2.4, 6) to [bend left=55]  (5.5, 2.2) to  [bend left=0] (0,0);
	\fill [pattern=north east lines, pattern color=lightgray] (0, 0) to (-2.4*1.2, 6*1.2) to [bend left=55]  (7, 0) to  [bend left=0] (0,0);
	\end{tikzpicture}    
&
\begin{tikzpicture}[scale=0.3]
\draw [thin] (-4.5, 0) -- (7, 0);
\draw [thin] (0, -1) -- (0, 8);

\node at (-2, 5) [inner sep=0pt, label=left:{\footnotesize $\sfz_1$}] {$\bullet$} ;
\node at (0, 1) [inner sep=0pt, label=left:{\footnotesize $\sfz_2$}] {$\bullet$};
\node at (5, 2) [inner sep=0pt, label=right:{\footnotesize $\sfz_3$}] {$\bullet$};
\node at (3, 4) [inner sep=0pt, label=right:{\footnotesize $\sfz_4$}] {$\bullet$};

\draw [dashed] (-2, 5) -- (0, 1) -- (5, 2) -- (3, 4) -- (-2, 5);
\fill [pattern=north west lines, pattern color=gray] (-2, 5) -- (0, 1) -- (5, 2) -- (3, 4) -- (-2, 5);
\fill [pattern=north east lines, pattern color=lightgray] (-2, 6.5) to (-2, 5) to (0, 1) to (7, 1) to [bend right=55]  (-2, 6.5);
\end{tikzpicture}
\\[-0.3cm]
(a) & (b)
\end{tabular}
\caption{Geometric representation of (a) the SHG Lemma \ref{Lemma: probability inequalities in SHG} and (b) the ME Lemma \ref{Lemma: probability inequalities in ME}.}
\label{figure: constraint characterization  of HG T-orders}
\end{figure}
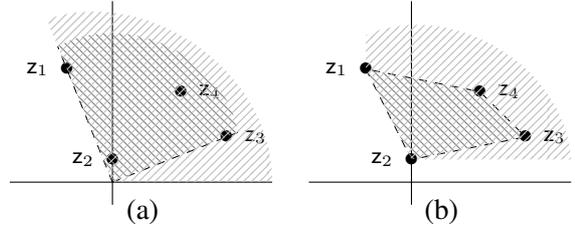

Lemma \ref{Lemma: probability inequalities in SHG} admits the following geometric interpretation, which will be used below. Suppose there are only $n=2$ constraints and $m=4$ losers $\sfz_i$. The difference vectors $\bfC(\sfx, \sfy, \sfz_i)$ which appear on the right hand side of (\ref{ex: Appendix; HG T-orders; equivalent condition}) can therefore be represented as the four black dots in Fig.~\ref{figure: constraint characterization  of HG T-orders}. The region $\{ \sum_{i=1}^m \lambda_i \bfC(\sfx, \sfy, \sfz_i) \,|\, \lambda_i \geq 0 \}$ is the {\em convex cone} generated by these four difference vectors $\bfC(\sfx, \sfy, \sfz_i)$, depicted in dark gray in Fig.~\ref{figure: constraint characterization  of HG T-orders}a. The region in light gray singles out the points which are at least as large as some point in this cone. Condition (\ref{ex: Appendix; HG T-orders; equivalent condition}) thus says that the difference vector $\bfC(\hatsfx, \hatsfy, \hatsfz)$ belongs to this light gray region.


\paragraph{ME}
The ME probability $\Prob_{\bfw}^{\mbox{\tiny ME}}(\sfy \,|\, \sfx)$ that a UR $\sfx$ is mapped to a SR $\sfy$ according to a nonnegative weight vector $\bfw$ is the exponential of the harmony $H_{\bfw}(\sfx, \sfy)$ of that mapping, normalized through a constant $Z = Z(\bfw, \sfx)$, namely $\Prob_{\bfw}^{\mbox{\tiny ME}}(\sfy \,|\, \sfx) = e^{H_{\bfw}(\sfx, \sfy)}/Z$. A\&M show that also in ME uniform probability inequalities can be characterized in terms of difference vectors, as stated by Lemma \ref{Lemma: probability inequalities in ME} below. This ME Lemma is analogous to the SHG Lemma \ref{Lemma: probability inequalities in SHG} above, but for two differences. The first difference is that condition (\ref{ex: Appendix; HG T-orders; equivalent condition}) is only necessary in ME while it is also sufficient in SHG. The second difference is that ME requires the {\em normalization condition} (\ref{equation: ME normalization condition}) on the coefficients $\lambda_i$.

\begin{Lemma} 		\label{Lemma: probability inequalities in ME}
Consider two mappings $(\sfx, \sfy)$ and $(\hatsfx, \hatsfy)$. Assume that the UR $\sfx$ comes with a finite number $m$ of loser candidates $\sfz_1, \dots, \sfz_m$ (besides the winner candidate $\sfy$). If the ME probability inequality $\Prob_{\bfw}^{\mbox{\tiny ME}}(\sfy \,|\, \sfx) \leq \Prob_{\bfw}^{\mbox{\tiny ME}}(\hatsfy \,|\, \hatsfx)$ holds uniformly for every choice of the nonnegative weight vector $\bfw$, then for every loser candidate $\hatsfz$ of the UR $\hatsfx$, there exist $m$ nonnegative coefficients $\lambda_1, \dots, \lambda_m \geq 0$ (one for each loser candidate $\sfz_1, \dots, \sfz_m$ of the UR $\sfx$) which add up to 1
\begin{equation}
\label{equation: ME normalization condition}
\lambda_1 + \dots + \lambda_m = 1
\end{equation}
and furthermore satisfy condition (\ref{ex: Appendix; HG T-orders; equivalent condition}). 
\hfill{$\Box$}
\end{Lemma}


The normalization condition (\ref{equation: ME normalization condition}) admits the following geometric interpretation.
As seen above, the region $\{ \sum_i \lambda_i \bfC(\sfx, \sfy, \sfz_i) \,|\, \lambda_i \geq 0 \}$ is the convex cone generated by the difference vectors $\bfC(\sfx, \sfy, \sfz_i)$, represented by the dark gray region in Fig.~\ref{figure: constraint characterization  of HG T-orders}a. The smaller region $\{ \sum_i \lambda_i \bfC(\sfx, \sfy, \sfz_i) \,|\, \lambda_i \geq 0, \fbox{$\sum_i \lambda_i = 1$} \}$, which differs for the (boxed) normalization condition (\ref{equation: ME normalization condition}) on the coefficients $\lambda_i$, is instead  the {\em convex hull} generated by the difference vectors $\bfC(\sfx, \sfy, \sfz_i)$, represented by the smaller dark gray region in Fig.~\ref{figure: constraint characterization  of HG T-orders}b. The effect of the normalization condition (\ref{equation: ME normalization condition}) is thus to shrink from the larger convex cone to the smaller convex hull. Finally, the region in light gray in Fig.~\ref{figure: constraint characterization  of HG T-orders}b singles out the points which are at least as large as some point in this convex hull. Lemma \ref{Lemma: probability inequalities in ME} thus requires the difference vector $\bfC(\hatsfx, \hatsfy, \hatsfz)$ to belong to this light gray region.

\section{ME has no equiprobable mappings}
\label{section: A new result on uniform probability identities in SHG and ME}

Lemmas \ref{Lemma: probability inequalities in SHG} and \ref{Lemma: probability inequalities in ME} say that ME differs from SHG because of the normalization condition (\ref{equation: ME normalization condition}). This apparently small technical difference has substantial phonological implications. Indeed, this Section shows that the normalization condition (\ref{equation: ME normalization condition}) makes the ME typology so rich that it can distinguish between any two mappings. In other words, equiprobability is impossible in ME. The reasoning is presented here informally, split up into three steps formalized in the final appendix.


\paragraph{Step 1}
Let us suppose that the two mappings $(\sfx, \sfy)$ and $(\hatsfx, \hatsfy)$ are equiprobable in ME, namely that the ME probability identity $\Prob_{\bfw}^{\mbox{\tiny ME}}(\sfy \,|\, \sfx) = \Prob_{\bfw}^{\mbox{\tiny ME}}(\hatsfy \,|\, \hatsfx)$ holds for every choice of the nonnegative weight vector $\bfw$. Let $\sfz_1, \dots, \sfz_m$ be the loser candidates of the UR $\sfx$. They define a light gray region as in Fig.~1b, namely the region of points which are at least as large as the points in the convex hull generated by the difference vectors $\bfC(\sfx, \sfy, \sfz_i)$. Let us denote this light gray region as $\mbox{\em LGR}^{\mbox{\tiny ME}}(\sfz_1, \dots, \sfz_m)$. Analogously, let $\hatsfz_1, \dots, \hatsfz_{\hatm}$ be the loser candidates of the other UR $\hatsfx$. They as well define the light gray region of points which are at least as large as the points in the convex hull generated by the difference vectors $\bfC(\hatsfx, \hatsfy, \hatsfz_j)$. Let us denote this light gray region as $\mbox{\em LGR}^{\mbox{\tiny ME}}(\hatsfz_1, \dots, \hatsfz_{\hatm})$.


The probability identity $\Prob_{\bfw}^{\mbox{\tiny ME}}(\sfy \,|\, \sfx) = \Prob_{\bfw}^{\mbox{\tiny ME}}(\hatsfy \,|\, \hatsfx)$ is equivalent to the two reverse inequalities $\Prob_{\bfw}^{\mbox{\tiny ME}}(\sfy \,|\, \sfx) \leq \Prob_{\bfw}^{\mbox{\tiny ME}}(\hatsfy \,|\, \hatsfx)$ and $\Prob_{\bfw}^{\mbox{\tiny ME}}(\sfy \,|\, \sfx) \geq \Prob_{\bfw}^{\mbox{\tiny ME}}(\hatsfy \,|\, \hatsfx)$. By lemma \ref{Lemma: probability inequalities in ME} above, the former inequality requires each difference vector $\bfC(\hatsfx, \hatsfy, \hatsfz_j)$ to belong to $\mbox{\em LGR}^{\mbox{\tiny ME}}(\sfz_1, \dots, \sfz_m)$. And the latter inequality requires each difference vector $\bfC(\sfx, \sfy, \sfz_i)$ to belong to $\mbox{\em LGR}^{\mbox{\tiny ME}}(\hatsfz_1, \dots, \hatsfz_{\hatm})$. A simple convexity argument deduces from these two facts the identity $\mbox{\em LGR}^{\mbox{\tiny ME}}(\sfz_1, \dots, \sfz_m) = \mbox{\em LGR}^{\mbox{\tiny ME}}(\hatsfz_1, \dots, \hatsfz_{\hatm})$ between the two light gray regions.


\paragraph{Step 2}
To proceed, let us suppose for concreteness that $m = 4$ and that the light gray region $\mbox{\em LGR}^{\mbox{\tiny ME}}(\sfz_1, \sfz_2, \sfz_3, \sfz_4)$ is the one plotted in light gray in Fig.~\ref{figure: constraint characterization  of HG T-orders}b. The difference vectors corresponding to the two losers $\sfz_1$ and $\sfz_2$ are {\em extreme points} (or {\em vertices}) of this light gray region. In the sense that they crucially contribute to shape it: if these two points were shifted even slightly in any direction, the corresponding light gray region would change. The identity between the two light gray regions established in step 1 thus entails that the two light gray regions share the same set of extreme points. In conclusion, the two difference vectors corresponding to losers $\sfz_1$ and $\sfz_2$ which are extreme points of the light gray region in figure Fig.~\ref{figure: constraint characterization  of HG T-orders}b must be shared by the two equiprobable mappings considered. Since these difference vectors are shared by the two equiprobable mappings, they can be ``peel off'' the two sides of the ME probability identity.


\paragraph{Step 3}
We are thus left with the difference vectors corresponding to the other two losers $\sfz_3$ and $\sfz_4$ in Fig.~\ref{figure: constraint characterization  of HG T-orders}b. These latter two vectors are not extreme points of the original light gray region but rather sit in the interior of the light gray region. Indeed, they can be shifted around without affecting the shape of the light gray region. Yet, once the two losers $\sfz_1$ and $\sfz_2$ have been ``peeled off'' at step 2, we can repeat the reasoning in steps 1 and 2 ignoring the two losers $\sfz_1$ and $\sfz_2$ and instead considering only the other two losers $\sfz_3$ and $\sfz_4$.

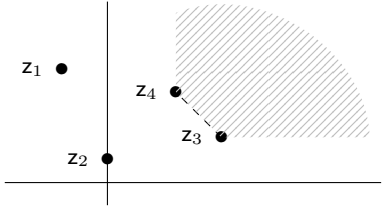
\begin{figure}[t]
\centering
\begin{tikzpicture}[scale=0.3]
\draw [thin] (-4.5, 0) -- (12, 0);
\draw [thin] (0, -1) -- (0, 8);

\node at (-2, 5) [inner sep=0pt, label=left:{\footnotesize $\sfz_1$}] {$\bullet$} ;
\node at (0, 1) [inner sep=0pt, label=left:{\footnotesize $\sfz_2$}] {$\bullet$};
\node at (5, 2) [inner sep=0pt, label=left:{\footnotesize $\sfz_3$}] {$\bullet$};
\node at (3, 4) [inner sep=0pt, label=left:{\footnotesize $\sfz_4$}] {$\bullet$};

\draw [dashed] (5, 2) -- (3, 4);
\fill [pattern=north east lines, pattern color=lightgray] (3, 7.5) to (3,4) to (5,2) to (11.5, 2) to [bend right=55]  (3,7.5);
\end{tikzpicture}
\caption{Steps 1-2 for the remaining losers $\sfz_3$ and $\sfz_4$.}
\label{figure: repeating the reasoning}
\end{figure}


Thus, we construct the convex hull of the difference vectors corresponding to just these two remaining losers $\sfz_3$ and $\sfz_4$. This convex hull is the segment which connects the two corresponding dots. Next, we construct the light gray region of points which are at least as large as some point in that segment, as depicted in Fig.~\ref{figure: repeating the reasoning}. Now the difference vectors corresponding to the two losers $\sfz_3$ and $\sfz_4$ are extreme points of the new light gray region. We can therefore repeat the reasoning in steps 1-2 and conclude that these two difference vectors as well must be shared by the two equiprobable mappings considered. And so on.


The reasoning informally sketched above leads to the following Proposition \ref{Proposition: equiprobable mappings in ME}, which is the first main result of this paper. It says that two mappings are equiprobable in ME if and only if they share all difference vectors. This entails in particular that the two mappings must have the same number of loser candidates. In other words, the ME typology is so rich that the only case where ME fails to come up with at least one weight vector which assigns different probabilities to the two mappings $(\sfx, \sfy)$ and $(\hatsfx, \hatsfy)$ is when the two mappings are the same mapping, in the sense that they are indistinguishable by the constraints, as they have the same difference vectors.\footnote{
To illustrate, suppose that the constraint set only consists of the two constraints \textsc{NoVoicedObstruent} and \textsc{Ident}(voice). The mappings $(\sfx, \sfy) = (\textsf{/\scriptsize mab/}, \textsf{\scriptsize [map]})$ and $(\hatsfx, \hatsfy) = (\textsf{\scriptsize /bam/}, \textsf{\scriptsize [pam]})$ will always have the same ME probability, because they and their losers have the same constraint violation profiles.
}
\begin{Proposition}
\label{Proposition: equiprobable mappings in ME}
Two mappings $(\sfx, \sfy)$ and $(\hatsfx, \hatsfy)$ are equiprobable in ME if and only if the corresponding sets of difference vectors coincide. 
\hfill{$\Box$}
\end{Proposition}

\section{SHG allows for equiprobable mappings}
\label{section: SHG allows for equiprobable mappings}

The preceding Section has shown that ME is so rich that it can distinguish between any two different mappings. Crucially, this typological richness is peculiar to ME, not intrinsic to probabilistic constraint-based phonology. In this section, we illustrate this point with the case of SHG. As in the preceding section, the discussion is kept informal. The formalization rests on the same convex geometric tools used for ME in the final appendix. The details are omitted here for reasons of space (see the longer version of this paper available on the authors' website).


Let us consider two mappings $(\sfx, \sfy)$ and $(\hatsfx, \hatsfy)$. Again, let $\sfz_1, \dots, \sfz_m$ be the loser candidates of the UR $\sfx$. They define a light gray region as in Fig.~\ref{figure: constraint characterization  of HG T-orders}a, namely the region of points which are at least as large as the points in the convex cone generated by the difference vectors $\bfC(\sfx, \sfy, \sfz_i)$. Let us denote this light gray region as $\mbox{\em LGR}^{\mbox{\tiny SHG}}(\sfz_1, \dots, \sfz_m)$. This region is different from (and larger than) the light gray region $\mbox{\em LGR}^{\mbox{\tiny ME}}(\sfz_1, \dots, \sfz_m)$ considered above for ME, because the latter ME region is restricted through the normalization condition (\ref{equation: ME normalization condition}) and therefore defined in terms of convex hulls rather than convex cones. Analogously, let $\hatsfz_1, \dots, \hatsfz_{\hatm}$ be the loser candidates of the other UR $\hatsfx$ and let $\mbox{\em LGR}^{\mbox{\tiny SHG}}(\hatsfz_1, \dots, \hatsfz_{\hatm})$ be the corresponding SHG light gray region.


Again as in the case of ME, Lemma \ref{Lemma: probability inequalities in SHG} says that the uniform SHG probability identity $\Prob_{\bfw}^{\mbox{\tiny SHG}}(\sfy \,|\, \sfx) = \Prob_{\bfw}^{\mbox{\tiny SHG}}(\hatsfy \,|\, \hatsfx)$ entails that the two SHG light gray regions coincide, namely that $\mbox{\em LGR}^{\mbox{\tiny SHG}}(\sfz_1, \dots, \sfz_m) = \mbox{\em LGR}^{\mbox{\tiny SHG}}(\hatsfz_1, \dots, \hatsfz_{\hatm})$. Yet, these SHG light gray regions have different geometric properties than the ME light gray regions. As a result, in the case of SHG the identity between the two light gray regions tells us much less about the difference vectors that generate them than in the case of ME.


To see that concretely, let us consider for instance the SHG light gray region in Fig.~\ref{figure: constraint characterization  of HG T-orders}a. The loser candidates $\sfz_2, \sfz_3$ and $\sfz_4$ have difference vectors which sit in the interior of this light gray region. These losers thus contribute nothing to shape the light gray region: their difference vectors can be shifted around without affecting the shape of the region. Identity of the light gray regions thus tells us nothing about identity of these difference vectors which sit in the interior.


Interestingly, the loser candidates whose difference vectors sit in the interior of the SHG light gray region can be characterized phonologically as those losers which are {\em HG redundant} given the rest of the losers. In the sense that, for every nonnegative weight vector $\bfw$, if the HG harmony of the winner $\sfy$ is larger than that of the nonredundant losers, then it is in particular larger than the harmony of the redundant losers. In other words, these redundant losers carry no interesting phonological content as they do not in any way affect the weight vectors consistent with the mapping $(\sfx, \sfy)$.


The case of the loser $\sfz_1$  in Fig.~\ref{figure: constraint characterization  of HG T-orders}a is instead different. Its difference vector sits on the border of the light gray region and therefore contributes to its shape. Yet, its position is not completely determined by the shape of the region. In fact, the shape of the region is not affected if this difference vector is slid closer to or further away from the origin. Equivalently, the shape of the region is not affected if the difference vector corresponding to the nonredundant loser $\sfz_1$ is rescaled by a nonnegative constant $\lambda \geq 0$. This means that the identity of the two SHG light gray regions does not entail identity of the difference vectors which generate them, not even for those difference vectors which sit on the boundary of the regions and therefore correspond to nonredundant losers. The identity of the two SHG light gray regions only entails that the difference vectors of the nonredundant losers are one the rescaling of the other. This informal reasoning leads to the following Proposition, which is our second main result.


\begin{Proposition}
\label{Proposition: equiprobable mappings in SHG}
Two mappings $(\sfx, \sfy)$ and $(\hatsfx, \hatsfy)$ are equiprobable in SHG if and only if each nonredundant difference vector $\bfC(\sfx, \sfy, \sfz_i)$ is a rescaling of some nonredundant difference vector $\bfC(\hatsfx, \hatsfy, \hatsfz_j)$, namely $\bfC(\sfx, \sfy, \sfz_i) = \lambda \bfC(\hatsfx, \hatsfy, \hatsfz_j)$ for some $\lambda \geq 0$; analogously, each nonredundant difference vector $\bfC(\hatsfx, \hatsfy, \hatsfz_j)$ is a rescaling of some nonredundant difference vector $\bfC(\sfx, \sfy, \sfz_i)$.
\hfill{$\Box$}
\end{Proposition}

Interestingly, this characterization of SHG equiprobability coincides with the characterization of equivalence in categorical HG obtained by A\&M. We conclude that two mappings are equiprobable in SHG (namely are always assigned the same probability) if and only if they are equivalent in categorical HG (namely no HG grammar succeeds on one but fails on the other).

\section{Equiprobability in Finnish stress}
\label{section: Equiprobable mappings in Finnish stress}

This section brings the preceding formal results to bear on Finnish word stress.


\begin{table}
\renewcommand{\tabcolsep}{0.1cm}\begin{tabular}{ll}
{\sc FtBin}		& Feet are disyllabic.\\
{\sc PkProm}		&No unstressed light syllables.\\
{\sc Align-L}		&All feet left.\\
{\sc *Rev} 		&No trochees with sonority reversal.\\
{\sc *Flat}		&No trochees with a flat sonority profile.\\
*H.X			&No stress next to a heavy syllable.\\
{\sc WSP}		&No unstressed heavy syllables.\\
{\sc WSP/VV}		& No unstressed heavies with long vowel.
\end{tabular}
\caption{Constraints for foot structure in Finnish nouns}
\label{finnstress-constraints}
\end{table}


\paragraph{The phonological system}
The basic generalizations about Finnish word stress can be stated as follows \cite{Carlson(1978),HansonKiparsky(1996),Elenbaas(1999),ElenbaasKager(1999),Karvonen(2005)}: (a) primary stress falls on the initial syllable; (b) secondary stress falls on every other syllable after that, (c) except that a light syllable is skipped if the syllable after that is heavy, unless the heavy syllable is final. Examples are \textsf{\small \'{i}l.moit.t\`{a}u.tu.m\`{i}.nen} `registering' and \textsf{\small \'{i}l.moit.t\`{a}u.tu.mi.s\`{e}s.ta} `from registering'.


\begin{table*}
\begin{tikzpicture}
\matrix [row sep = -0.5cm, column sep = -0.05cm] {
\node (block one) [align = center, draw] { \scriptsize\textsf{(j, (kon.sul)(taa.ti.o)ja)} 0.5\%\\[-0.15cm] \scriptsize\textsf{(i, (kom.mu)(ni.ke.o)ja)} 0.3\%\\[-0.15cm] \scriptsize\textsf{ (g, (o.pe)(raa.ti.o)ja)} 0.0\%\\[-0.15cm] \scriptsize\textsf{ (h, (al.le)(go.ri.o)ja)} 0.0\%};	
&
\node {$\leq$};
&
\node (block three) [align = center, draw, red] {\scriptsize\textsf{(c, (sym.po)(si.u.me)ja)} 98.6\%\\[-0.15cm] \scriptsize\textsf{(e, (po.ly)(a.mi.de)ja)} 95.7\%\\[-0.15cm] \scriptsize\textsf{(f, (in.ku)(naa.be.le)ja)} 9.5\%\\[-0.15cm] \scriptsize\textsf{(d, (lii.rum)(laa.ru.me)ja)} 18.6\%};
&
\node {$\leq$};
&
\node (block five) [align = center, draw] {\scriptsize\textsf{(b, (pro.pa)(gan.dis.te)ja) 100\%} \\[-0.15cm] \scriptsize\textsf{(a, (ak.va)(rel.lis.te)ja)} 100\%};
&
&
\\
&&&&&&
\node (block seven) [align = center, draw] {\scriptsize\textsf{ (k, (ter.mos)(taat.te)ja)} 100\%\\[-0.15cm] \scriptsize\textsf{(l, (mar.ga)(rii.ne)ja)} 100\%\\[-0.15cm] \scriptsize\textsf{(m, (af.fri)(kaat.to)ja)} 99.7\%};
\\
\node (block two) [align = center, draw] {\scriptsize\textsf{(b, (pro.pa)(gan.dis)(tei.ta))} 0.0\% \\[-0.15cm] \scriptsize\textsf{(a, (ak.va)(rel.lis)(tei.ta))} 0.0\%};	
&
\node {$\leq$};
&
\node (block four) [align = center, draw, red] {\scriptsize\textsf{(e, (po.ly)(a.mi)(dei.ta))} 4.3\% \\[-0.15cm] \scriptsize\textsf{(d, (lii.rum)(laa.ru)(mei.ta))} 81.4\%\\[-0.15cm] \scriptsize\textsf{(c, (sym.po)(si.u)(mei.ta))} 1.4\%\\[-0.15cm] \scriptsize\textsf{(f, (in.ku)(naa.be)(lei.ta))} 90.5\%};
&
\node {$\leq$};
&
\node (block six) [align = center, draw] {\scriptsize\textsf{ (h, (al.le)(go.ri)(oi.ta))} 100\% \\[-0.15cm] \scriptsize\textsf{(i, (kom.mu)(ni.ke)(oi.ta))} 99.7\%\\[-0.15cm] \scriptsize\textsf{(j, (kon.sul)(taa.ti)(oi.ta))} 99.5\%\\[-0.15cm] \scriptsize\textsf{(g, (o.pe)(raa.ti)(oi.ta))} 100\%};
&
\node {\phantom{$\;$}};
\\
};
\node at (4.7, 1) {\begin{rotate}{-45}{$\leq$}\end{rotate}};
\node at (4.7, -1.2) {\begin{rotate}{45}{$\leq$}\end{rotate}};
\end{tikzpicture}
\caption{Seven blocks of equiprobable mappings predicted by SHG}
\label{table: equiprobable SHG mappings}
\end{table*}
\begin{table*}
\begin{tikzpicture}
\matrix [row sep = 0.2cm, column sep = -0.05cm] {
\node (block four) [align = center, draw] {\scriptsize\textsf{(c, (sym.po)(si.u)(mei.ta))} 1.4\%};
&
\node {$\leq$};
&
\node (block four) [align = center, draw] {\scriptsize\textsf{(e, (po.ly)(a.mi)(dei.ta))} 4.3\%};
&
\node {$\leq$};
&
\node (block four) [align = center, draw] {\scriptsize\textsf{(d, (lii.rum)(laa.ru)(mei.ta))} 81.4\%}; 
&
\node {$\leq$};
&
\node (block four) [align = center, draw] {\scriptsize\textsf{(f, (in.ku)(naa.be)(lei.ta))} 90.5\%};
\\
\node (block four) [align = center, draw] {\scriptsize\textsf{(c, (sym.po)(si.u.me)ja)} 98.6\%};
&
\node {$\leq$};
&
\node (block four) [align = center, draw] {\scriptsize\textsf{(e, (po.ly)(a.mi.de)ja)} 95.7\%}; 
&
\node {$\leq$};
&
\node (block four) [align = center, draw] {\scriptsize\textsf{(d, (lii.rum)(laa.ru.me)ja)} 18.6\%};
&
\node {$\leq$};
&
\node (block four) [align = center, draw] {\scriptsize\textsf{(f, (in.ku)(naa.be.le)ja)} 9.5\%};
\\
};
\end{tikzpicture}
\caption{SHG's two red blocks are split into two chains of uniform inequalities in ME}
\label{table: ME inequalities}
\end{table*}


However, the skipping clause turns out to be a coarse approximation of the actual facts. Skipping is sometimes optional and we find variable stress in cases like \textsf{\small pr\'{o}.fes.so.r\`{i}l.la}$\sim$\textsf{\small pr\'{o}.fes.s\`{o}.ril.la} `professor-{\sc ade}', where the basic rule fails at the second variant. This optional pattern turns out to depend on two additional conditions that affect the outcome in a gradient manner \cite{Anttila(2012)}: (a) low vowels (\textsf{\small /a, \"{a}, o, \"{o}/}) attract stress and high vowels (\textsf{\small /e, i, u, y/}) repel stress; (b) stress is avoided next to a heavy syllable.\footnote{
The categories ``low" and ``high" are morphophonemic, not phonetic. In Finnish, low vowels alternate morphophonologically with rounded mid vowels (\textsf{\scriptsize a $\sim$ o, \"{a} $\sim$ \"{o}}) and the unrounded high vowel alternates with the unrounded mid vowel (\textsf{\scriptsize i $\sim$ e}). For this reason we consider \textsf{\scriptsize o, \"{o}} low and \textsf{\scriptsize e} high.
} 


In addition to native speaker intuitions about syllable prominence, empirical support for these soft conditions can be obtained from the optional rule of {\em Stop Deletion} \cite{KeyserKiparsky(1984)} which deletes singleton stops in extrametrical syllables \cite{Anttila(2012)}. In particular, the \textsf{\small /t/} in the partitive suffix \textsf{\small /-tA/} is deleted vs.~retained depending on the location of secondary stress feet. Given the input \textsf{\small /professori-i-tA/} `professor-{\sc pl-par}' we have two possible foot structures: (\textsf{\small pr\'{o}.fes.so\/})(\textsf{\small r\`{e}i.ta\/}) where \textsf{\small /t/} falls inside a foot and is retained vs.~(\textsf{\small pr\'{o}.fes\/})(\textsf{\small s\`{o}.re\/})\textsf{\small ja\/} where \textsf{\small /t/} falls outside a foot and is deleted. The metrical free variation is thus reflected in segmental free variation. This provides a valuable diagnostic for foot structure, especially because the segmental variation is present even in the written standard language readily available in large quantities.


The constraints necessary for deriving the foot structure in Finnish nouns are shown in Table \ref{finnstress-constraints}. These constraints were applied to 48 types of partitive plural nouns, systematically varying stem length, syllable weight, and vowel sonority. All in all, the test set contains 4 types of three-syllable stems, 12 types of 4-syllable stems, and 32 types 5-syllable stems (stem types are briefly denoted as ``(a), (b), \dots'' in what follows).


\paragraph{SHG}
We computed the uniform probability inequalities predicted by SHG for this Finnish stress test case using $\mathbb{C}$o$\mathbb{G}$e$\mathbb{T}$o \cite{MagriAnttila(2019CoGeTo)}, a suite of tools for studying constraint-based typologies of categorical and probabilistic phonological grammars based on their underlying rich convex geometry. The key observation is that SHG predicts seven blocks of equiprobable mappings, shown in Table \ref{table: equiprobable SHG mappings}. These blocks are furthermore organized into two chains of uniform probability inequalities. The predicted probabilities increase from left to right. The symbol ``$\leq$" between two boxes means that the candidates in the box on the left are predicted to have a probability at most as large as the candidates in the box on the right.

To evaluate the empirical accuracy of the equiprobabilities predicted by SHG, we examined Finnish \textsf{\small /t/}-deletion in a corpus of approximately 9 million nouns (tokens) harvested from Finnish internet sites on April 12, 2005. The percentages reported in Table \ref{table: equiprobable SHG mappings} represent the token frequency of \textsf{\small /t/}-retention vs.~\textsf{\small /t/}-deletion variants for each phonologically distinct stem type. The corpus data are consistent with the equiprobability prediction in five out of the seven blocks, namely those in black. These blocks turn out to be empirically nearly categorical, with almost all stems undergoing either \textsf{\small /t/}-deletion or \textsf{\small /t/}-retention, consistently with the equiprobability prediction.


However, the two red blocks in Table \ref{table: equiprobable SHG mappings} bundle together the stem types (c)-(f) despite them showing rather different empirical frequencies, providing {\em prima facie} evidence against SHG's equiprobability prediction. The stem types are illustrated by \textsf{\small /symposiumi/} `symposium', \textsf{\small /polyamidi/} `polyamide', \textsf{\small /liirumlaarumi/} `nonsense', and \textsf{\small /inkunaabeli/} `incunable'. The stems differ in the weight and quality of the preantepenultimate and antepenultimate syllables (heavy vs.~light, $[+\mbox{low}]$ vs.~$[-\mbox{low}]$), which results in constraint violation differences, yet HG predicts that all four should undergo \textsf{\small /t/}-deletion/retention at identical rates. In order to reconcile SHG's equiprobability predicitions with corpus frequencies, we make the following observations. First, the difference between types (d) \textsf{\small /liirumlaarumi/} and (f) \textsf{\small /inkunaabeli/} is not statistically significant ($\chi^2$ = 2.9849, {\em df\/} = 1, {\em p\/} = 0.08404). Second, type (c) contains only two stems: \textsf{\small /symposiumi/} `symposium' and \textsf{\small /imperiumi/} `empire', both potentially syllabifiable as four-syllable stems, e.g., \textsf{\small im.pe.ri.u.mi\/} $\sim$ \textsf{\small im.pe.riu.mi\/} \cite{AnttilaShapiro(2017)}, which is consistent with their unexpectedly high \textsf{\small /t/}-deletion rate. This leaves us with type (e) \textsf{\small /polyamidi/} `polyamide' (N = 69), again with an unexpectedly high deletion rate for which we have no plausible explanation. We conclude that by and large our Finnish corpus data support SHG's equiprobability predictions.

\paragraph{ME}

One might wonder whether ME with its ability to make fine-grained distinctions might actually offer a more principled solution to the difficulties just discussed. This turns out {\em not} to be the case. On the retention side, ME predicts the uniform probability inequalities in the top row of Table \ref{table: ME inequalities}. For example, the retention probability of \textsf{\small /polyamidi/} is predicted to be at most as high as that of \textsf{\small /liirumlaarumi/}, no matter the choice of the weight vector. That seems initially promising: these inequalities are in fact exactly what we observe in the data. Puzzlingly, on the deletion side, ME reverses the probabilities, yielding the uniform probability inequalities in the bottom row of Table \ref{table: ME inequalities}. For example, the deletion probability of \textsf{\small /polyamidi/} is predicted to be at most as high as that of \textsf{\small /liirumlaarumi/}. This is exactly the opposite of what we observe in the data. We submit there is simply no way to reconcile ME's predictions with the corpus data. Such counterintuitive probability reversals appear in other blocks as well. 


\section{Summary and conclusions}

We have shown that ME predicts typologies so rich that ME grammars can distinguish between any two different mappings and therefore admit no equiprobable mappings (Proposition 1). This richness does not extend to other implementations of probabilistic constraint-based phonology, such as SHG (Proposition 2), revealing a fundamental difference between the two frameworks. 

We have then applied these results to the test case of Finnish word stress. Our corpus data provide preliminary evidence in favor of SHG's equiprobability predictions. In the two blocks where SHG appeared to run into problems, ME did not help refine the analysis empirically, but instead split the SHG equiprobable stem types apart in a counterintuitive fashion. Our study thus provides some preliminary empirical support in favor of SHG, which permits equiprobable mappings, against ME, which does not.

\section*{Acknowledgements}

For very useful discussion, we would like to thank the participants of the workshop {\em Analyzing typological structure: from categorical to probabilistic phonology}
(Stanford, September 2018; \href{https://sites.google.com/site/analyzingtypologicalstructure/}{\underline{website}}). The research reported in this paper has been supported by a Collaborative Projects Grant from the France-Stanford Center for Interdisciplinary Studies (project title: {\em The Mathematics of Language Universals}) as well as by a JCJC grant from the Agence Nationale de la Recherche (project title: {\em The mathematics of segmental phonotactics}).

\appendix
\section{Proof of Proposition \ref{Proposition: equiprobable mappings in ME}}
\label{Appendix: ME T-orders have no cycles}

We write $\bfc_i$ and $\hatbfc_j$ as shorthands for the difference vectors $\bfC(\sfx, \sfy, \sfz_i)$ and $\bfC(\hatsfx, \hatsfy, \hatsfz_j)$ corresponding to the losers $\sfz_i$ and $\hatsfz_j$. 
The ME probability inequality $\Prob_{\bfw}^{\mbox{\tiny ME}}(\sfx, \sfy) = \Prob_{\bfw}^{\mbox{\tiny ME}}(\hatsfx, \hatsfy)$ can be made explicit as in (\ref{equation: Appendix; ME probability identity}) through some elementary manipulations. As usual, $\bfa^{\sf T} \bfb$ denotes the scalar product of $\bfa$ and $\bfb$.
\begin{equation}
\label{equation: Appendix; ME probability identity}
\mbox{$\sum_{i=1}^m e^{\bfw^{\sf T} \bfc_i} = \sum_{j=1}^{\hatm} e^{\bfw^{\sf T} \hatbfc_j}$}
\end{equation}
Once the ME probability identity $\Prob_{\bfw}^{\mbox{\tiny ME}}(\sfx, \sfy) = \Prob_{\bfw}^{\mbox{\tiny ME}}(\hatsfx, \hatsfy)$ is made explicit as in (\ref{equation: Appendix; ME probability identity}), it is obvious that it holds uniformly for every weight vector $\bfw$ when the two sets of difference vectors coincide, namely $\{ \bfc_1, \dots, \bfc_m \} = \{ \hatbfc_1, \dots, \hatbfc_{\hatm} \}$. To complete the proof of Proposition \ref{Proposition: equiprobable mappings in ME}, we thus only have to prove the reverse. We split the proof into three steps, corresponding to those in Section \ref{section: A new result on uniform probability identities in SHG and ME}.


\smallskip\noindent\textbf{Step 1.}
We start from the assumption that the ME probability identity $\Prob_{\bfw}^{\mbox{\tiny ME}}(\sfx, \sfy) = \Prob_{\bfw}^{\mbox{\tiny ME}}(\hatsfx, \hatsfy)$ holds uniformly. This means in particular that the probability inequality $\Prob_{\bfw}^{\mbox{\tiny ME}}(\sfx, \sfy) \leq \Prob_{\bfw}^{\mbox{\tiny ME}}(\hatsfx, \hatsfy)$ holds uniformly. The necessary condition for this uniform ME inequality provided by Proposition \ref{Lemma: probability inequalities in ME} can be rewritten as the inclusion (\ref{ex: T-orders in ME; cycles; first step; one}). As usual, $\real_+$ is the set of nonnegative real numbers and $A+B = \{ \bfa + \bfb \,|\, \bfa \in A, \bfb \in B \}$ is the vector sum of two sets $A$ and $B$ of $\real^n$. The region on the right hand side of (\ref{ex: T-orders in ME; cycles; first step; one}) is the light gray region in Fig.~\ref{equation: ME normalization condition}.b. 
\begin{exe}
\ex
\label{ex: T-orders in ME; cycles; first step; one}
$\displaystyle \{ \hatbfc_1, \dots, \hatbfc_{\hatm} \} \subseteq conv( \bfc_1, \dots, \bfc_m) + \real_+^n$
\end{exe}
The set $conv( \bfc_1, \dots, \bfc_m) + \real_+^n$ on the right hand side of (\ref{ex: T-orders in ME; cycles; first step; one}) is convex because the two sets $conv( \bfc_1, \dots, \bfc_m)$ and $\real_+^n$ are both convex and the sum of two convex sets is convex \cite[Section 2.3.2]{BoydVandenberghe(2004)}. The inclusion (\ref{ex: T-orders in ME; cycles; first step; one}) thus extends from the points $\hatbfc_1, \dots, \hatbfc_{\hatm}$ to their convex hull $conv(\hatbfc_1, \dots, \hatbfc_{\hatm})$, yielding the inclusion $conv(\hatbfc_1, \dots, \hatbfc_{\hatm}) \subseteq conv( \bfc_1, \dots, \bfc_m) + \real_+^n$. Finally, by adding $\real_+^n$ at both sides, the latter inclusion entails $conv(\hatbfc_1, \dots, \hatbfc_{\hatm}) + \real_+^n \subseteq conv( \bfc_1, \dots, \bfc_m) + \real_+^n$. Analogously, the reverse probability inequality $\Prob_{\bfw}^{\mbox{\tiny ME}}(\hatsfx, \hatsfy) \leq \Prob_{\bfw}^{\mbox{\tiny ME}}(\sfx, \sfy)$ requires the reverse inclusion $conv(\bfc_1, \dots, \bfc_m) + \real_+^n \subseteq  conv( \hatbfc_1, \dots, \hatbfc_{\hatm}) + \real_+^n$, yielding (\ref{ex: T-orders in ME; cycles; second step}). 
\begin{equation}	
\label{ex: T-orders in ME; cycles; second step}
\!\!\!\!\!\underbrace{conv( \bfc_1\dots\bfc_m) \!+\! \real_+^n}_{P} \!=\! \underbrace{conv(\hatbfc_1\dots \hatbfc_{\hatm}) \!+\! \real_+^n}_{\widehat{P}}
\end{equation}

\noindent\textbf{Step 2.}
This identity (\ref{ex: T-orders in ME; cycles; second step}) says in particular that the two sets $P$ and $\widehat{P}$ on its left and right hand side have the same set of extreme points, namely $ext(P) = ext(\widehat{P})$. The set $ext(P)$ of extreme points of the set $P$ is nonempty. In fact, a set which is closed, convex, nonempty, and does not contain a line admits at least an extreme point \cite[Proposition 2.1.2]{Bertsekas(2009)}. Indeed, $P$ is closed, because $conv(\bfc_1, . . . , \bfc_m)$ is compact, $\real_+^n$ is closed, and the sum of a compact set with a closed set is closed \cite[Section 1.3]{Bertsekas(2009)}. Furthermore, $P$ is convex, because $conv(\bfc_1, . . . , \bfc_m)$ and $\real_+^n$ are both convex and the sum of two convex sets is convex. Finally, $P$ is obviously nonempty and it does not contain a line.


The set $ext(P)$ of extreme points of the set $P$ is a subset of the set of difference vectors $\{ \bfc_1, \dots, \bfc_m \}$. In fact, the set of extreme points of the finitely generated polyhedron $conv(\bfc_1, \dots, \bfc_m)$ is a subset of $\{ \bfc_1, \dots, \bfc_m \}$ (by the Krein-Milman theorem). The set of extreme points of the pointed cone $\real_+^n$ only consists of the zero vector $\mathbf{0}$. And the set $ext(A+B)$ of extreme points of the vector sum $A+B$ of any two polyhedra $A$ and $B$ is a subset of the vector sum $ext(A) + ext(B)$ of the two sets $ext(A)$ and $ext(B)$ of extreme points of $A$ and $B$, namely $ext(A+B) \subseteq ext(A) + ext(B)$ \cite[exercise 2.22]{BertsimasTsitsiklis(1997)}. Analogously, the set $ext(\widehat{P})$ of extreme points of the set $\widehat{P}$ is a nonempty subset of the set $\{ \hatbfc_1, \dots, \hatbfc_m \}$.


In conclusion, the two sets of difference vectors $\{ \bfc_1, \dots, \bfc_m \}$ and $\{ \hatbfc_1, \dots, \hatbfc_m \}$ share the vectors in the nonempty set $\Omega = ext(P) = ext(\widehat{P})$. Without loss of generality, we assume that these shared vectors are those corresponding to the first $h \geq 1$ losers, so that $\{ \bfc_1, \dots, \bfc_m \} = \Omega \cup \{ \bfc_{h+1}, \dots, \bfc_m \}$ and $\{ \hatbfc_1, \dots, \hatbfc_m \}	= \Omega \cup \{ \hatbfc_{h+1}, \dots, \hatbfc_{\hatm} \}$.


\smallskip\noindent\textbf{Step 3.}
The terms on the left and the right hand side of the ME probability identity (\ref{equation: Appendix; ME probability identity}) which correspond to the shared difference vectors in $\Omega$ cancel out. The ME probability identity thus reduces to $\sum_{i=h+1}^m e^{\bfw^{\sf T} \bfc_i} = \sum_{j=h+1}^{\hatm} e^{\bfw^{\sf T} \hatbfc_j} $, where the sums start at $h+1$ rather than at $1$. The claim follows by iterating the reasoning above, starting from the latter simplified ME probability identity.

\bibliography{PhonologyBibliography}
\bibliographystyle{acl_natbib_nourl}
\end{document}